# Word Segmentation as Graph Partition

Yuanhao Liu[1] and Sheng Yu[2,3*]

[1]*Department of Statistics, University of Michigan, Ann Arbor, MI, USA*

[2]*Center for Statistical Science, Tsinghua University, Beijing, China*

[3]*Department of Industrial Engineering, Tsinghua University, Beijing, China*

[*]Correspondence to:
Sheng Yu, Weiqinglou Rm 209, Tsinghua University, Beijing, 100084, China
Email: syu@tsinghua.edu.cn

## Abstract

We propose a new approach to the Chinese word segmentation problem that considers the sentence as an undirected graph, whose nodes are the characters. One can use various techniques to compute the edge weights that measure the connection strength between characters. Spectral graph partition algorithms are used to group the characters and achieve word segmentation. We follow the graph partition approach and design several unsupervised algorithms, and we show their inspiring segmentation results on two corpora: (1) electronic health records in Chinese, and (2) benchmark data from the Second International Chinese Word Segmentation Bakeoff.

## 1. Introduction

The Chinese text is written as a sequence of characters without delimiters to mark the beginning and the end of words. While human readers can figure out the words from the text with minimum effort, the absence of explicit delimiters creates one more difficulty for computers to analyze the language. Therefore, word segmentation – correctly breaking down a given sentence into a sequence of words – is the first step in almost any Chinese natural language processing (NLP) task.

Word segmentation has been a long time study given its importance in Chinese NLP. The lack of a complete dictionary for Chinese (new words are created rapidly, especially on the internet; a huge amount of informal words are not recorded by any dictionary), the overlap of possible words ("新任职务" potentially includes "新任", "任职", and "职务"), and varying granularity ("外交部长" can be seen as a single word or as two words "外交" and "部长") without a clear rule make Chinese word segmentation a challenging task. Among methods benchmarked with annotated corpora, the most accurate have been the supervised ones, which generally treat word segmentation as a sequential classification problem. However, unsupervised methods that do not rely on annotated corpora are particularly of interest, because when we need domain-specific NLP, which could be NLP for a specific industry or for historical text that has a different language pattern from that of today's, segmenters pre-trained with annotated text from the general domain (such as news text) will not work satisfactorily, and there is no annotated corpus available to train a segmenter for the target text, due to the intensive labor required for the annotation.

Unlike human readers who can rely on his/her vast knowledge and experience in the language, unsupervised methods generally resort to information theory or statistical principles, such as placing a word boundary according to the conditional entropy [1] or choosing the segmentation plan that maximizes the likelihood [2–4]. These approaches work to some extent, but their accuracy is limited because word segmentation has complex underlying rules that cannot be described solely by probability and statistics. It is also hard to improve on top of the existing approaches, because they do not allow feature engineering, which is crucial to many highly accurate machine learning algorithms.

In this paper, we propose a new approach to Chinese word segmentation by solving it as a graph partition problem. We consider a sentence as an undirected graph, where each character in the sentence corresponds to a node, and the edge weight $w_{ij}$ is the connection strength between nodes $i$ and $j$. Graph partitioning is a well-studied problem in mathematics that seeks to partition a graph into small components (words, in our case), so that the total weight of edges running across components is small. The key to a satisfying partition is in the proper design of the connection strength matrix $W$, which is analogous to the kernel in support vector machines that represents similarity, and in its design one can leverage not only information theory and statistical principles, but also one's knowledge and experience in the language or the target text, making it ultimately flexible.

The main contributions of this paper are as follows.

1. We propose the graph partition approach to the word segmentation problem. Characters in a sentence are treated as nodes of a graph, with matrix $W$ characterizing the connection strengths between each pair of nodes. One can use well-established graph partition algorithms for word segmentation.
2. We develop an unsupervised algorithm for word segmentation in electronic health records (EHR). The transition probability-based design of the matrix $W$ allows effective segmentation of words and medical named entities. A tunable parameter allows adjustment of the granularity. Given the fact that there is not a comprehensive medical terminology in Chinese, this unsupervised dictionary-independent algorithm will be highly useful for Chinese EHR processing.
3. We develop another two algorithms and test them on benchmark data. One algorithm utilizes a generic dictionary and the other utilizes words in the training data to modify the connection strength matrix $W$. While the F-scores of the two algorithms are lower than that of supervised algorithms, properties demonstrated by the segmentation results can be preferable in many applications.

## 2. Related Work

Lots of methods have been proposed for Chinese word segmentation. Dictionary-based methods use heuristic rules to segment a sentence according to recorded words in a dictionary. The most successful dictionary-based methods are variants of the maximum matching (MM) algorithm, including forward MM, backward MM and Bi-directional MM [5–9]. These methods can be highly accurate when the words in the sentence are covered by the dictionary. However, dictionary coverage is very limited for both general and domain texts, and its compilation cannot catch up with the creation of new words. Thus, in practice, the effectiveness of dictionary-based methods is limited.

Supervised methods are more intelligent than dictionary-based methods in that they can learn from an annotated corpus and do not require a complete dictionary to work well. The word segmentation problem are commonly considered as a sequential tagging problem, where each character is assigned a label, such as a B label for the beginning of a word and an N label for non-beginnings. Hidden Markov models [10], maximum entropy [11,12] and conditional random field [13–15] are all effective models for solving this problem. In recent years, with the rise of deep learning, many neural-based models have also been proposed. Zheng et al. used deep learning to achieve Chinese word segmentation and part-of-speech tagging together [16]. Pei et al. constructed a max-margin tensor neural network to model complicated interactions between tags and context characters [17]. Chen et al. extended the long-short-term-memory (LSTM) model to explicitly model previous information in memory cells [18]. Overall, supervised models achieve the highest F-scores on benchmark data. However, the performance of models trained with the benchmark data (most of which are news text) can drop quickly on other texts such as those from specialized domains.

Although not as accurate as supervised models in benchmark tests, unsupervised models are widely useful because they can adapt to various texts, especially those from specialized domains, without the need of expert annotation. The unsupervised models generally rely on information theory or statistical principles. Sproat and Shih proposed a recursive procedure that used mutual information to identify two-character words from a sentence [19]. Sun and Shen extended the work of Sproat and Shih by adding the difference of t-scores and complex rules to achieve better segmentation [20]. Several studies employed the maximum likelihood principle and used the expectation-maximization (EM) algorithm to segment the sentence [2–4]. Jin and Tanaka-Ishii proposed to place a word boundary whenever the conditional entropy started to increase or was greater than a threshold [1]. Our proposed approach of treating word segmentation as a graph partition problem has a similarity with the maximum likelihood approaches in that they are both global optimizations, which is a theoretical advantage over the other methods that only make local decisions. The advantage of our approach over the maximum likelihood approaches is the flexibility of design: we are free to use anything from information theory, probability, statistics, to vocabularies, heuristics, and prior knowledge of the target text to design the character connection strength and improve the satisfaction of segmentation.

## 3. The Graph Partition Approach for Word Segmentation

In this section, we quickly review the spectral graph partition theory and connect it to the word segmentation problem. For further reading of the spectral graph partition theory and proofs, see Chung [21] and von Luxburg [22].

Consider an undirected graph $G$, whose nodes $\{v_1, \dots, v_n\}$ represent the $n$ characters of a given sentence in corresponding order. $W$ is an $n \times n$ symmetric matrix that the segmentation algorithm designer needs to provide, where entry $w_{ij}$ ($w_{ij} \geq 0$ for all $i, j$) is the edge weight between nodes $v_i$ and $v_j$ and characterizes the connection strength between the $i$th and the $j$th characters. A strong connection with large $w_{ij}$ indicates that characters $i$ and $j$ are likely to be in the same word. For instance, on a scale from 0 to 1, if "麒麟" appears in a sentence, the connection strength between the two characters should ideally be 1, because they always form a word when they appear together and neither of them binds with other characters. Note that $w_{ij}$ is not necessarily determined solely by the two characters, and should be computed given the other characters in the sentence. For instance, $w_{ij}$ between "天" and "门" in "天 X 门" intuitively should be large if X is "安", and should be much smaller if X is any other character. Weight $w_{ij} = 0$ means that there is no edge between nodes $v_i$ and $v_j$, but they can still be connected by a path via other nodes and be in the same word.

Spectral partition algorithms are a class of graph partition algorithms that utilize the properties of the graph Laplacian matrix $L$, which has several alternative forms. In this paper, we consider the unnormalized form

$$L = D - W,$$

and the symmetric normalized form

$$L_{sym} = I - D^{-1/2} W D^{-1/2},$$

where $I$ is the $n \times n$ identity matrix, $D$ is the diagonal matrix of node degrees, i.e., $d_i = \sum_{j=1}^{n} w_{ij}$. It can be shown that $L$ and $L_{sym}$ are positive semidefinite, with at least one eigenvalue equal to 0.

In the ideal case, assume that each word in the sentence forms a connected component of $G$. That is, every two characters in the same word are connected by a path whose intermediate nodes are all characters of that word, and characters from different words are not connected by any path. In this special case, the Laplacian matrix has the following property.

**Proposition 1** *If graph $G$ has connected components denoted by $A_1, \dots, A_k$, and $1_{A_i}$ is the 0/1 vector that indicates whether each node is in $A_i$, then the multiplicity of the eigenvalue 0 for both $L$ and $L_{sym}$ equals to $k$. For $L$, the eigenspace of 0 is spanned by the vectors $1_{A_i}$, $i = 1, \dots, k$. For $L_{sym}$, the eigenspace of 0 is spanned by $D^{1/2} 1_{A_i}$, $i = 1, \dots, k$.*

One can use a spectral decomposition algorithm to compute the $k$ eigenvectors $u_1, \dots, u_k$ of $L$ corresponding to 0, the smallest eigenvalue. By **Proposition 1**, $u_1, \dots, u_k$ are linear combinations of $1_{A_1}, \dots, 1_{A_k}$:

$$U = [u_1 \dots, u_k] = [1_{A_1} \dots 1_{A_k}] C,$$

where $C$ is a $k \times k$ matrix. It is easy to see that if nodes $v_i$ and $v_j$ are in the same connected component (word), then the $i$th and $j$th rows of $U$ are identical. Conversely, if the nodes are from different connected components, then the corresponding rows of $U$ are different as well. For the normalized Laplacian matrix $L_{sym}$, since the eigenspace of 0 is spanned by $D^{1/2} 1_{A_i}$, $i = 1, \dots, k$, the same property holds if we normalize $U$ by row.

In practice, we do not expect that each word in the sentence forms a connected component by itself, or the word segmentation problem would be trivial. However, by proper design of the connection strength matrix $W$, we expect that the edge weights running across words to be much smaller than those running within words. Therefore, by the perturbation theory, if there are $k$ words in the sentence, we expect that the rows of $U$ (normalize the rows for $L_{sym}$) to form $k$ easily distinguishable clusters, and we can identify them with simple clustering algorithms, such as the k-means. Since we do not know the value of $k$, we can determine it as the number of eigenvalues of $L$ or $L_{sym}$ within a certain range from 0. The effectiveness of a word segmentation algorithm therefore is strongly related to the design of the connection strength matrix $W$.

The choice of the graph Laplacian matrix also affects the segmentation quality. The two forms $L$ and $L_{sym}$ correspond to two close but different optimization criteria. Clustering using the unnormalized Laplacian matrix $L$ is the continuous approximation to the *ratio cut* problem [23]:

$$\min_{A_1,\dots,A_k} \sum_{i=1}^{k} \frac{W(A_i, A_i^c)}{|A_i|},$$

where $W(A,B) = \sum_{v_i \in A, v_j \in B} w_{ij}$ is the total weight from subset $A$ to subset $B$, and $|A|$ is the number of nodes in $A$. Clustering using the normalized Laplacian matrix $L_{sym}$ is the continuous approximation to the *normalized cut* problem [24]:

$$\min_{A_1,\dots,A_k} \sum_{i=1}^{k} \frac{W(A_i, A_i^c)}{\mathrm{vol}(A_i)},$$

where $\mathrm{vol}(A) = \sum_{v_i \in A} w_{ij}$ is the total weight of all edges attached to the nodes in $A$. The two optimization criteria have different effects, and the normalized cut is more likely to segment a sentence into shorter words than the ratio cut. The choice of the Laplacian matrix therefore depends on the application.

## 4. Experiments

In this section, we demonstrate the use of the graph partition method for word segmentation in two scenarios. The first experiment is word segmentation for electronic health records, where we design an algorithm without using any dictionary. The second experiment is a test on the benchmark data from the Second International Chinese Word Segmentation Bakeoff (ICWB2) [25].

### 4.1 Word segmentation for electronic health records

Electronic health records (EHR) contain lots of free text that require natural language processing. One of the most important tasks in EHR analysis is named entity recognition – identifying the medical terms and map them to pre-defined concepts of an ontology. This task is relatively easy in English, because the terms, including synonyms and abbreviations, are quite completely recorded in ontologies such as the Unified Medical Language System (UMLS) [26]. However, for Chinese, a similar comprehensive terminology does not exist. Therefore, one would heavily rely on word segmentation to identify the terms. Since clinical text is very specialized, existing word segmenters trained with general text such as news corpora do not work on clinical text satisfactorily. Annotated clinical corpora for training supervised algorithms are not available, either. Here we provide an unsupervised algorithm following the graph partition approach, without using a dictionary. The EHR text we used were in-patient EHR from a pneumology department, measuring about 100 MB in size.

**Algorithm 1**

```
s = input character sequence
n = s.length
# construct W
W = diag(1, n) # identity matrix
for (i in 1:(n-1)):
    W[i,i+1] = max(P(s[i+1]|s[i]), P(s[i+1]|s[i-1],s[i]),
    P(s[i]|s[i+1],s[i+2])) * sd_count_bi(s[i:i+1])
    if (s[i] or s[i+1] in weaken_set_1)
        W[i,i+1] = W[i,i+1]/4
    if (s[i] or s[i+1] in weaken_set_2)
        W[i,i+1] = W[i,i+1]/80
for (i in 1:(n-2)):
    W[i,i+2] = P(s[i+1],s[i+2]|s[i]) * sd_count_tri(s[i:i+2])
    if (s[i], s[i+1] or s[i+2] in weaken_set_2)
        W[i,i+2] = 0
Copy the upper triangle of W to its lower triangle.
# partitioning
D = diag(rowSums(W))
L = D - W # the unnormalized form
eig = eigen(L) # eigen decomposition with decreasing eigenvalues
k = sum(eig$values <= eig.cut)
U = eig$vectors[,(n-k+1):n]
clusters = kmeans(U, k)
return s[clusters]
```

Some boundary conditions are omitted in **Algorithm 1** for conciseness. `P(s[i+1]|s[i])` is the transition probability estimated from the corpus that the next character is `s[i+1]` given the current character `s[i]`. In this experiment, transition probabilities from/to non-Chinese characters such as English letters and digits were set to 0. `P(s[i+1]|s[i-1],s[i])` is the probability that the next character is `s[i+1]` given the current character `s[i]` and the previous character `s[i-1]`. `P(s[i]|s[i+1],s[i+2])` is the probability that the current character is `s[i]` given the next two characters `s[i+1]` and `s[i+2]`. The function `sd_count_bi(bigram)` is the logarithm of the count of `bigram` in the corpus divided by the standard deviation of the logarithm of the count of a random bigram. Similarly, `sd_count_tri(trigram)` is the standardized logarithm of the count of `trigram`. The conditional weakening of entries of `W` checks if the bigram or trigram contains a character that

tends to form a word by itself. In our experiment, `weaken_set_1` includes 和是在对中与将要地以为有, and `weaken_set_2` includes 了的无及等行不. We know from our experience of the language that the characters in `set_2` are almost always used as a one-character word in EHR while those in `set_1` may still form words with other characters. In this experiment, both the characters in the two sets and the weakening factors 4 and 80 are chosen subjectively and empirically work well – we use this to demonstrate that we can freely mix data-driven methods and empirical knowledge for the design of $W$, which we see as a strength of the proposed approach. The `eig.cut` is a user specified positive parameter that determines the granularity. The initial centers of the k-means clustering can be specified by the k-means++ algorithm [27] or picked evenly from the rows of `U`. One may need to add a random noise (e.g., from $N(0,0.001)$) to the entries of `U` to prevent identical centers.

Table 1: Examples of word segmentation in EHR under various granularities.

| |
|---|
| input = "患者无咳嗽、咳痰，无咯血，无发热等不适症状", `eig.cut` = 0.15<br>result = 患者 无 咳嗽 、 咳痰 ， 无 咯血 ， 无 发热 等 不适 症状 |
| input = "明确诊断原发性右肺下叶周围型鳞癌", `eig.cut` = 0.15<br>result = 明确诊断 原发性右肺下叶周围型鳞癌 |
| input = "明确诊断原发性右肺下叶周围型鳞癌", `eig.cut` = 2<br>result = 明确 诊断 原发性 右肺下叶 周围型 鳞癌 |
| input = "冠状动脉粥样硬化性心脏病", `eig.cut` = 0.15<br>result = 冠状动脉粥样硬化性心脏病 |
| input = "冠状动脉粥样硬化性心脏病", `eig.cut` = 2<br>result = 冠状动脉 粥样硬化性 心脏病 |
| input = "甲状腺功能异常", `eig.cut` = 0.1<br>result = 甲状腺功能异常 |
| input = "甲状腺功能异常", `eig.cut` = 0.15<br>result = 甲状腺功能 异常 |
| input = "甲状腺功能异常", `eig.cut` = 2<br>result = 甲状腺 功能 异常 |
| input = "并给予吉西他滨、顺铂药物治疗", `eig.cut` = 0.15<br>result = 并给予 吉西他滨 、 顺铂 药物治疗 |
| input = "并给予吉西他滨、顺铂药物治疗", `eig.cut` = 2<br>result = 并 给予 吉西他滨 、 顺铂 药物 治疗 |
| input = "腹部脐下正中线可见手术疤痕", `eig.cut` = 0.15<br>result = 腹部脐下正中线 可见 手术疤痕 |
| input = "腹部脐下正中线可见手术疤痕", `eig.cut` = 1<br>result = 腹部 脐下 正中线 可见 手术 疤痕 |

**Table 1** shows representative segmentation results of sentences from the EHR. It is remarkable to see that with merely transition probabilities the graph partition approach easily identifies drug names such as "吉西他滨" (gemcitabine) and "顺铂" (cisplatin) as well as other medical terms without using any dictionary. We can also tune the level of granularity with the `eig.cut` parameter. A small `eig.cut` value gives a segmentation result that is great for identifying named entities, such as "冠状动脉粥样硬化性心脏病" (coronary atherosclerotic cardiopathy).

As we increase the value of `eig.cut`, the segmentation becomes finer; for example, “冠状动脉粥样硬化性心脏病” is segmented as “冠状动脉” (the coronary artery), “粥样硬化性” (atherosclerotic), and “心脏病” (cardiopathy). Another great example is that “甲状腺功能异常” (thyroid function abnormal) can be segmented as a single term, as “甲状腺功能” (thyroid function) “异常” (abnormal), or as “甲状腺” (thyroid) “功能” (function) “异常” (abnormal) by tuning `eig.cut` and all the segmentations make sense. However, while segmentation results under a finer granularity is closer to what many linguists define as “words” (as described in the segmentation guidelines of many annotated training sets), from the EHR analysis point of view, a coarser granularity is more useful. For instance, both long terms “冠状动脉粥样硬化性心脏病” and “甲状腺功能异常” are clinical phenotypes and well recognized medical named entities (UMLS concepts C0010054 and C0476414, respectively). Either way, the user is able to adjust the granularity according to his/her needs.

Prior knowledge of a problem can be critically important when there is not enough annotated data for supervised learning. Being able to freely modify the connection strength using prior knowledge, like the conditional weakening in **Algorithm 1**, is a key advantage of the proposed method over methods purely based on information theory or statistical principles. The characters in `weaken_set` 1 and 2 can bind relatively strongly with other characters in terms of transition probability. Having the prior knowledge that these characters generally should form words by themselves allows us to create rules to artificially weaken their connections to other characters. The rules in **Algorithm 1** are still crude. For example, when weakening the connection strengths involving the character “及” (commonly used alone in general text that means “and”), a better rule should exclude the bigrams “闻及”, “触及”, “未及”, which are all common words in EHR.

The transition probability-based algorithm utilizes the fact that the character transition probability within a word is much larger than that across words. However, the text in EHR is sometimes based on a template, so the transition from word to word can lose the randomness necessary for the algorithm to work. For example, the phrase “主任医师查房记录” (attending physician’s record of ward round) commonly appear as fixed in the EHR, and the algorithm can have difficulty segmenting the words in this case.

### 4.2 Word segmentation on benchmark data

We were also interested in knowing how the graph partition approach would perform on benchmark data. We tested two algorithms on the annotated test set given by Peking University in ICWB2. The two algorithms were largely similar to **Algorithm 1**. The transition probabilities were computed using news articles of People’s Daily from 1994-01-01, to 2001-02-12. Since the annotation in the test set had a very small granularity, to achieve a similar effect, we removed the `W[i,i+2]` entries in the connection strength matrix, and we chose to use the normalized graph Laplacian matrix instead of the unnormalized one. In addition, instead of the conditional weakening in **Algorithm 1**, in **Algorithm 2** we used the dictionary “现代汉语常用词表” (Common Vocabulary of Modern Chinese) published by The Commercial Press to modify the connection strength matrix $W$. The vocabulary contains 56064 words ranked by frequency from high to low. An entry `W[i,i+1]` was multiplied by 20 if the corresponding bigram `s[i:i+1]` was part of a word with a rank less than 25000. If `W[i,i+1]` was not strengthened and if `s[i:i+1]` contained a character $A$ from the single character word set "的在地和向是上中下不有对并了与将还但就要以为也而又于", then `W[i,i+1]` was divided by $\max(20, \ln(10^6/rank(A)))$. We used a uniform `eig.cut` value 0.00035. Some post-processing rules were added, such as concatenating numbers and dates. The recall, precision, and F-score were 0.86, 0.84, and 0.85, respectively.

**Algorithm 3** differed from **Algorithm 2** in that it used the words and their frequencies from the training set instead of a dictionary to modify $W$. An entry `W[i,i+1]` was multiplied by 20 if the corresponding bigram `s[i:i+1]` was part of a word in the training set. If `W[i,i+1]` contained a character $A$ that was a single-character word in the training set, then `W[i,i+1]` was divided by $\max(1, count(A)/250)$. So strictly speaking, **Algorithm 3** was not unsupervised, but "minimally supervised". We set `eig.cut` = 0.001. The recall, precision, and F-score were 0.90, 0.87, and 0.89, respectively.

Table 2: Examples of word segmentation on the benchmark data. Important differences are marked in bold.

| |
|---|
| Gold-standard: 在 这个 激动人心 的 时刻 ， 我 很 高兴 通过 中国 国际 **广播 电台** 、 中央 人民 **广播 电台** 和 中央 电视台 ， 向 全国 各族 人民 ， 向 香港 特别 行政区 同胞 、 澳门 **特别 行政区** 同胞 和 台湾 同胞 、 海外 侨胞 ， 向 **世界 各国** 的 朋友 们 ， 致以 新 世纪 **第一 个** 新年 的 祝贺 ！<br>Algorithm 3: 在 这个 激动人心 的 时刻 ， 我 很 高兴 通过 中国 国际 **广播 电台** 、 中央 人民 **广播 电台** 和 中央 电视台 ， 向 全国 各族 人民 ， 向 香港 特别 行政区 同胞 、 澳门 **特别 行政区** 同胞 和 台湾 同胞 、 海外 侨胞 ， 向 **世界各国** 的 朋友 们 ， 致以 新 世纪 **第一个** 新年 的 祝贺 ！<br>Algorithm 2: 在 这个 激动人心 的 时刻 ， 我 很 高兴 通过 中国 国际 **广播电台** 、 中央 人民 **广播电台** 和 中央 电视台 ， 向 全国 **各族人民** ， 向 香港 **特别行政区** 同胞 、 澳门 **特别行政区** 同胞 和 台湾 同胞 、 海外 侨胞 ， 向 **世界 各国** 的 朋友 们 ， 致以 新 世纪 **第一个** 新 年 的 祝贺 ！ |
| Gold-standard: 演出 开始 前 ， **尉 健行** 、 **李 岚清** 会见 了 参加 演出 的 主要 演员 。<br>Algorithm 3: 演出 开始 前 ， **尉 健行** 、 **李 岚清** 会见 了 参加 演出 的 主要 演员 。<br>Algorithm 2: 演出 开始 前 ， **尉健行** 、 **李岚清** 会见 了 参加 演出 的 主要 演员 。 |
| Gold-standard: 显然 ， 进入 21 世纪 ， **中国 人** 会 生活 得 更为 忙碌 ， 思想 更为 解放 ， 环境 更为 宽松 和谐 ， **中华 文化** 会 更为 灿烂 。<br>Algorithm 3: 显然 ， 进入 21 世纪 ， **中国人** 会 生活 得 更为 忙碌 ， 思想 更为 解放 ， 环境 更为 宽松 和 谐 ， **中华文化** 会 更为 灿烂 。<br>Algorithm 2: 显然 ， 进入 21 世纪 ， **中国人** 会 生活 得 更为 忙碌 ， 思想 更为 解放 ， 环境 更为 宽松 和谐 ， **中华文化** 会 更为 灿烂 。 |
| Gold-standard: **降水 概率** 20％<br>Algorithm 3: **降水 概率** 20％<br>Algorithm 2: **降水概率** 20％ |

The dictionary-based modification of $W$ allowed the segmentation result to largely imitate the segmentation criteria used by the gold-standard. Changing the dictionary from a general vocabulary to words appeared in the training data easily improved the F-score by 4 percent points, showing the importance of the training data in getting a high score. In fact, while we focus on unsupervised word segmentation in this paper, the graph partition approach does not preclude supervised learning of $W$, which will surely further raise the F-score on benchmark data. On the other hand, word segmentation is not a game of getting the highest F-score on benchmark data, and the algorithm design should be driven by the needs of the applications (Gao et al. 2005). Table 2 shows a few representative examples where the segmentation results differed from the gold-standard. We see that the two transition probability-based algorithms (especially Algorithm 2, which used less information from the benchmark data) tended to merge "words" that strongly bound as a concept, entity, or common expression together. For example, "中国人" (Chinese), instead of "中国" (China) "人" (man); "尉健行" (the full name), instead of "尉" (the family name) "健行" (the given name).[1] Thus, the segmentation by the two algorithms can be more favorable than the gold-standard in many applications.

## 5. Conclusion

In this paper, we proposed the graph partition approach for the classical problem of Chinese word segmentation. The key to effective segmentation is the appropriate design of the connection strength matrix, then one can use well-established graph partition algorithms to segment the sentence. The computation of the connection strength matrix allows great flexibility, including using prior knowledge of the target text, which is a key advantage of the graph partition approach over existing unsupervised algorithms based solely on information or statistical principles. For demonstration of the methodology, we developed simple transition probability-based unsupervised algorithms for word segmentation in EHR and benchmark data, respectively, and obtained inspiring results.

On the other hand, the study of the graph partition approach to word segmentation is still primitive. A more thorough study of various possible designs of the connection strength and the corresponding Laplacian matrix for different kinds of text is absolutely warranted. From the EHR data, we also found that the optimal granularity (the `eig.cut` parameter) may not be uniform across the text for the need of the application. Thus, how to choose granularity adaptively is another important problem worth further studying.

## Acknowledgement

The authors thank the following people for their help in data collection and processing:
Qiuyang Yin (yinqy15@mails.tsinghua.edu.cn), Tsinghua University, Beijing, China;
Shengxuan Luo (luosx14@mails.tsinghua.edu.cn), Tsinghua University, Beijing, China;
Ji Li (liji15@mails.tsinghua.edu.cn), Tsinghua University, Beijing, China.

---

[1] Unlike in English, a Chinese person is never referred to by his/her family name alone, nor by his/her given name alone if it has only one character. So, regardless of the linguistic definition of a "word" used to create the gold-standard, separating Chinese family name and given name apart is pointless from a practical point of view.